\documentclass[conference]{IEEEtran}
\IEEEoverridecommandlockouts
\usepackage{url}
\usepackage{subfig}
\usepackage{cite}
\usepackage{amsmath,amssymb,amsfonts}
\usepackage{algorithmic}
\usepackage{graphicx}
\usepackage{textcomp}
\usepackage{adjustbox}
\usepackage{graphicx}
\usepackage{xcolor}
\usepackage[textsize=tiny,textwidth=1.4cm]{todonotes}
\usepackage{mathtools,lipsum, nccmath,cuted}
\usepackage{siunitx}
\usepackage{textcomp}

\usepackage{comment}
\usepackage{makecell}
\usepackage{algorithm}
\usepackage{algorithmic}
\usepackage{gensymb}
\usepackage{ulem}
\usepackage{colortbl}
\usepackage{hyperref}
\usepackage{enumitem}
\definecolor{Gray}{gray}{0.925}
\usepackage[textsize=tiny]{todonotes}

\graphicspath{ {figures/} }
\def\BibTeX{{\rm B\kern-.05em{\sc i\kern-.025em b}\kern-.08em
    T\kern-.1667em\lower.7ex\hbox{E}\kern-.125emX}}
    
\usepackage{pgfplots}
\usepgfplotslibrary{colorbrewer}
\pgfplotsset{compat = 1.14, cycle list/Set1-8} 
\usetikzlibrary{pgfplots.statistics, pgfplots.colorbrewer,pgfplots.dateplot,
        shadows,
        matrix} 

\usepackage{pgfplotstable}
\usepackage{filecontents}
\usepackage{graphicx}
\pgfplotsset{compat=1.14}

\definecolor{blueLine}{RGB}{57,106,177}
\definecolor{blueFill}{RGB}{114,147,203}
\definecolor{redLine}{RGB}{204,37,41}
\definecolor{greenline}{RGB}{0,250,0}
\definecolor{blackLine}{RGB}{0,0,0}
\definecolor{goldLine}{RGB}{160,82,45}

\usepackage[font=footnotesize]{caption}
\begin{document}

\title{Deep Learning based Finite Element Analysis (FEA) surrogate for sub-sea pressure vessel}

\author{\IEEEauthorblockN{Harsh Vardhan,Janos Sztipanovits}
\IEEEauthorblockA{
\textit{Department of computer science}\\
\textit{Vanderbilt University} \\ Nashville, USA
}
}

\maketitle
\section{\textbf{\textit{ABSTRACT}}}
\textbf{During the design process of an autonomous underwater vehicle (AUV), the pressure vessel has a critical role. The pressure vessel contains dry electronics, power sources, and other sensors that can not be flooded. A traditional design approach for a pressure vessel design involves running multiple Finite Element Analysis (FEA) based simulations and optimizing the design to find the best suitable design which meets the requirement. Running these FEAs are computationally very costly for any optimization process and it becomes difficult to run even hundreds of evaluation. In such a case, a better approach is the surrogate design with the goal of replacing FEA-based prediction with some learning-based regressor. Once the surrogate is trained for a class of problem, then the learned response surface can be used to analyze the stress effect without running the FEA for that class of problem.  The challenge of creating a surrogate for a class of problems is data generation. Since the process is computationally costly, it is not possible to densely sample the design space and the learning response surface on sparse data set becomes difficult. During experimentation, we observed that a Deep Learning-based surrogate outperforms other regression models on such sparse data.  
In the present work, we are utilizing the Deep Learning-based model to replace the costly finite element analysis-based simulation process. By creating the surrogate we speed up the prediction on the other design much faster than direct Finite element Analysis. We also compared our DL-based surrogate with other classical Machine Learning (ML) based regression models( random forest and Gradient Boost regressor). We observed on the sparser data, the DL-based surrogate performs much better than other regression models. Code of training and data can be found at  \url{https://github.com/vardhah/FEAsurrogate}.
}  
\subsection*{Keywords}
\textbf{\textit{deep learning, finite element analysis, surrogate modeling, random forest, gradient boost, sub-sea pressure hull}}

\section{Introduction}
\label{sec:introduction}

Design problems in the engineering and scientific community are an ongoing challenge and in most cases, these design problems include the complex relationship between design parameters. Design space exploration is a process to understand the effect of design variables on the performance metrics and analyze the complex relationship between these variables on the performance metrics.  
The process usually involves successive sampling in the design space and analyzing the performance metric to find a promising design point or subspace which satisfies all the constraints and meets the necessary performance requirement. Learning these nonlinear hyper-planes about the relationship between input design variables and performance metrics is the  role of a human design engineer. For this purpose, the design engineer evaluates/simulates the design space at various samples and tries to create a response surface in his mind. This response surface is further used in the future to find the design for a given performance metric.  Generally, Humans are not very good at learning a nonlinear complex manifold. However, Machine Learning models have a proven record of automatic feature extraction and learning highly non-linear behavior. Learning these hyperplanes using these learning models  can assist human designers to find a good design in the highly complex nonlinear relationships.
However creating an ML-based nonlinear manifold for a class of problems involves multiple issues: first, generating enough samples is difficult in a high dimension design problem. Second, the approximation error of trained models converges
to acceptable error only with a large amount of data, which is often difficult and expensive to obtain for complex simulation models, such as FEA.
Raissi et al\cite{raissi2019physics}proposed physics-informed ML approaches by encoding known physics-based invariant and constraints to the learning process which make learning possible with less data.  However deriving any Physics-based ML for general discretized models, such as finite element models has not got any practical success. 

In the present work, our goal is to design an ML-based surrogate for the FEA process for sub-sea pressure vessels. For a given design space, the aim is to create a data-driven learning-based surrogate model for the simulation process. The benefit of creating a surrogate is parallel faster evaluation of other design points by a trained model, and it will speed up the whole design decision process.  
These surrogates are then can be used for the outer optimization loop. By automatically designing models from data the paradigm of model-driven engineering is shifting from rather than utilizing the existing models and running optimization in an outer loop\cite{vardhan2019modeling} to first designing a model from generated data and then running optimization on learning models.

In conclusion, we want to share two results:
\begin{itemize}
   \item With a very handful of generated data, our learning model is able to predict test data with high accuracy of approx 92\%.
    \item We tested the accuracy of our deep learning architecture-based trained model in comparison to other classical regression models -random forest-based regression model and gradient boost-based regression model. We observed significant-higher accuracy of our model in comparison to both models.
\end{itemize}

The remainder of this paper is organized as follows: Section 2 formally defines a problem statement. In Section 3, we present our experiments and will evaluate the results. Section 4 discusses related work. We will conclude in Section 5 with a brief discussion of future work.

\section{Problem Formulation}
\label{sec:problem}
 
\begin{figure}[h!]
        \centering
        \captionsetup{justification=centering}
        \includegraphics[width=0.5\textwidth]{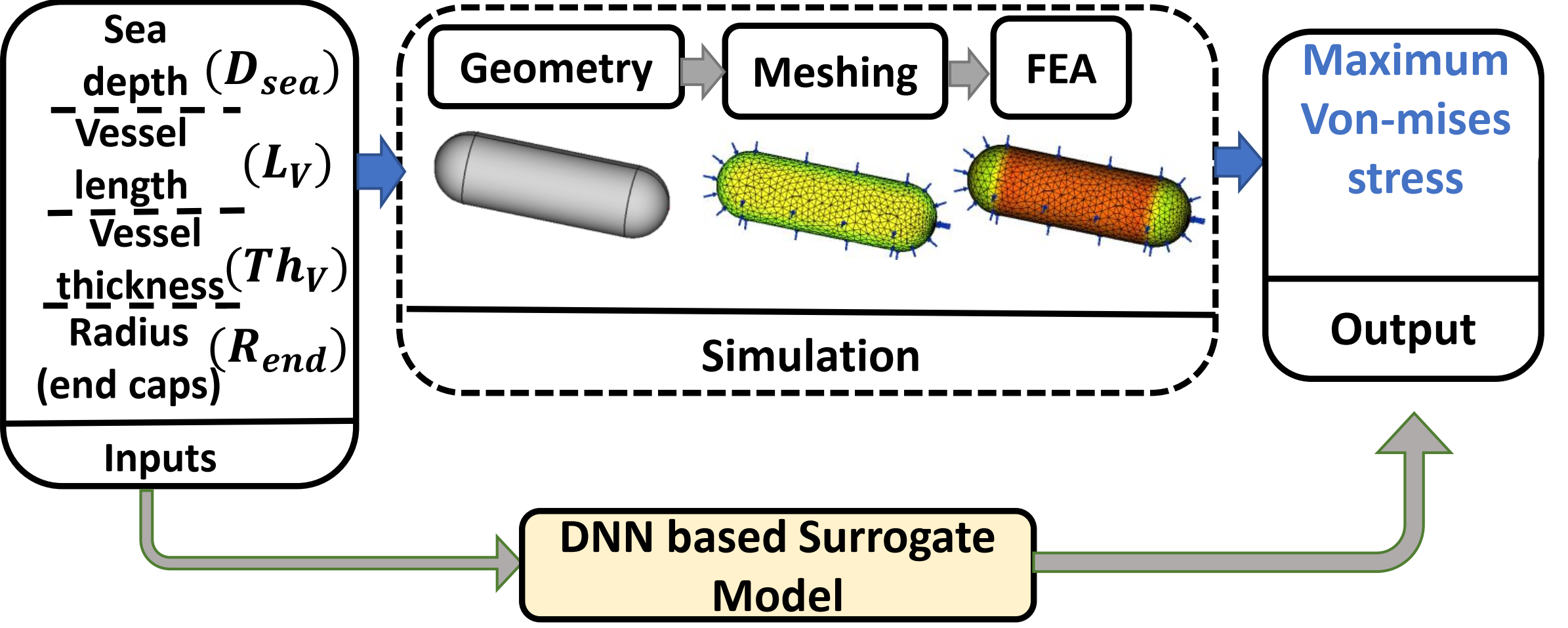}
        \caption{FEA Analysis and Surrogate modeling process}
        \label{fig:smp}
\end{figure}

 The pressure vessel is a critical component in modern-day Unmanned Underwater vehicle, Remotely operated Vehicles, etc. It is designed to withstand the sub-sea hydro-static pressure conditions while remaining watertight and contains power sources, electronic and other sensors that cannot be flooded. For this purpose, FEA-based simulation is conducted to measure the maximum induced stress in a design subject to the subsea pressurized environment. The simulation consists of multiple steps - CAD modeling, body meshing, and numerical solution of FEA to get the stress distribution. 
 The measured maximum static stress is compared with the yield strength of the material to check the integrity of the pressure vessel during operation. This process is repeated until an optimal design that can withstand the sub-sea pressure is found. In this work, our goal is to learn a surrogate model for this entire simulation process for a large class of problems. Since each simulation is computationally very costly ($\approx 202$ sec/simulation on a 20 core CPU), it is not possible to run many simulations and generate the dense data.   The motivation for learning a surrogate is the ability to interpolate or predict the numerical simulators' output at a very low cost using trainable models. Once trained the surrogate can replace the numerical simulators for outer loop optimization operations. The expected benefit is the ability to design the pressure vessel on a given requirement at a very small computational time for a range of problems. 
 The class of problem is defined by the design space (which are parameters that can be changed for changing the design and environmental conditions).   
 For learning a surrogate, we first create a design space. Our design space consists of the maximum depth of the sea, which decides the maximum applied sub-sea pressure on the pressure vessel. Most pressure vessels have a circular cylindrical shell with two hemisphere end caps configuration, justified by its benefits as balanced structural integrity, homeostatic balance,  and better internal volume. This cylindrical pressure vessel is parameterized by 3 parameters (length of cylinder, radius of end-cap, thickness of vessel). Accordingly, the design space consist of 4 variable space ($D_{sea},L_v,Th_v,R_{end}$) (refer figure \ref{fig:smp}). In this paper, we are interested in designing a range of pressure vessels that can work between range 0-6000 meters of sea depth. The material used to design the vessel is Aluminium alloy (Al6061-T6) which is the most commonly used material for pressure vessel design. 

The training process of the ML model starts with the collection of simulated data($D_{simulated}$). The simulation involves the design of CAD geometry based on design parameters, the meshing of CAD designs, application of desired hydrostatic pressure on an external surface of design, and then numerical solution using finite element analysis. Once the FEA simulation converges, the maximum Von-mises stress is measured(refer figure \ref{fig:smp}). After collection of training data, we split it into train data ($D_{train}$) and test data($D_{test}$).
Let $DS$ be a bounded subspace of $\mathbb{R}^d$, i.e., $DS \subset \mathbb{R}^d $ and it is represented by an set of $n$ samples. $d$ is the number of input parameters that describes the relevant system  properties. The training data is $D_{train}=(X,Y^*)$, where $X$ and  $Y^*$ are the collection of all $x_i $ (input samples to the simulator) and $y_i^*$ (corresponding simulation's output).  The role of a simulator is to map the input space to the solution space. i.e. $\textrm{Sim}:X \rightarrow \mathbb{Y} $.   
Accordingly, the surrogate modeling problem is the conditioning of a parametric model ($h$) on $D_{train}$ to find its parameters ($\theta$) such that:  
$ h(x, \theta|D_{train}) =y, \; \forall \,x \in DS : y \approx y^* $ where $y^*$ represents the ground truth.

\section{Experimentation setup}
\label{sec:experiments}
\subsection{Our Deep Neural Networks}
A deep neural network is an interconnected network of elementary computation units called neurons. Multilayered neural networks are one of the most powerful learning algorithms, in which these neurons are organized in layers. Neurons in the previous layer are connected to neurons at the next layer. Our NN architecture is a multilayered NN with 9 layers. Our network has four major constituents- fully connected layer (linear+ReLU), dropout layer, skip connections, and output layer. The input layer is the concatenation of the normalized vector of design variables and sub-sea pressure ($[D_{sea}, L_V, Th_V, R_{end}]$). The fully connected layer (linear+ReLU) consists of linear weights and biases with the commonly used nonlinear activation function ReLU (Rectified Linear Unit \cite{xu2015empirical}) defined by $f(x)=max(x,0)$. All the neurons in previous layer are connected to every neuron in current layer. Two dropout layers are added in between these fully connected layer for regularisation and to prevent over-fitting with a dropout factor of $0.2$.  The output of the FEA simulation is a scalar value $s \in \mathbb{R}$, so our choice of output layer is a linear fully connected layer. Last important feature is the skip connection, which is a residual network added for smooth learning and to handle vanishing gradient or exploding gradient\cite{he2016deep}. 

\begin{figure}[h!]
        \centering
        \captionsetup{justification=centering}
        \includegraphics[width=0.4\textwidth]{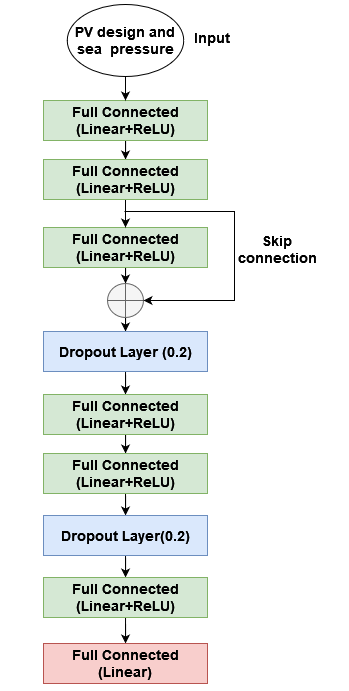}
        \caption{Deep Neural Network architecture}
        \label{fig:dnn}
    \end{figure} 

The overall architecture is represented in figure \ref{fig:dnn}. The network has given the input  in a batch of n with each sample vector size is $4$. The network has total 9 layers with six fully connected nonlinear layers, two dropout layers (dropout factor=0.2), and one fully connected linear output layer. The number of parameters in our architecture is 22,993. A large network(22,993 parameters) in comparison with the small data size(8000 training data points) raises concerns about over-fitting. The drop-out layer and cross-validation-based early stopping are used to counter this in our regression model.

\subsection{Training setup}
\begin{figure}[h!]
        \centering
        \captionsetup{justification=centering}
        \includegraphics[width=0.4\textwidth]{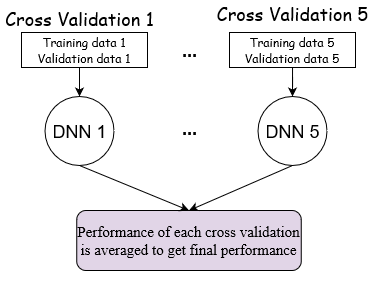}
        \caption{Ensemble of learning models}
        \label{fig:ensemble}
    \end{figure}

We divide the collected data(11311 data points) randomly into a training sample containing 8000 FEA evaluations and a testing sample containing the remaining 3311 evaluations. Due to the unavailability of large data, to increase the performance of prediction, we trained a family of base deep learning networks as it has been seen to be more accurate than individual classifiers\cite{sagi2018ensemble}. For this purpose, we used 5-fold cross-validation and trained one base deep neural network on each fold dataset. We further divided each fold dataset into the training dataset (90\%) and the validation dataset(10\%). The prediction of the individual trained networks was then averaged out to obtain the final values of the maximum Von-mises stress on a given design(refer figure \ref{fig:ensemble}). After experimenting with different loss functions and learning rates, we found that L1 loss performed better than other losses with a learning rate of 0.001. We used Adam\cite{kingma2014adam} optimizer with Xavier initialization for training the model. 

\subsection{Other ML models:}
Apart from our deep learning-based learning model, we also trained the decision trees-based model to predict the maximum Vonmises stress. Most popular decision trees based predictive model random forest and Gradient boost regressor is trained which are considered the best classification and regression models and
most popular predictive analytic techniques among practitioners, due to being relatively
straightforward to build and understand, as well as handling both nominal and continuous
inputs \cite{breiman2017classification}. We trained 100 different decision trees and tuned their depth, number of trees, and split criteria to get the most accurate model.
   
\subsection{Evaluation metrics} To assess the quality of prediction, we used following common statistics: 
\begin{enumerate}
    \item average \textbf{residual}, $\Delta Z= (V_{truth}- V_{predicted})/V_{truth}$ per sample 
    \item the \textbf{accuracy}, percentage of number of samples whose residual is within acceptable error of 10\% i.e $|\Delta Z| <0.10$, 
    \item $\eta$: the number of \textbf{outlier},  $|\Delta Z| >0.50$, number of samples who which prediction has high error.  
    \item the \textbf{standard deviation},($\sigma$) of prediction.
\end{enumerate}
Here, $V_{truth}$ and $V_{predicted}$ are the real label and the predicted outcome of the trained regression models. For comparison purposes, apart from our model we also trained other regression models i.e. Random forest regression model and Gradient Boosting for the regression model. Models have been developed and trained using SKlearn,
PyTorch, and XGBoost (extreme gradient boosting) python libraries. Code and data can be found here.

\section{Results on Model performance metrics}
The model selection process is to find the best balance between model fit and complexity. For regression modeling, the quantitative performance metrics are residual/mean absolute
error (MAE) and root mean squared error (RMSE)/ standard deviation. We also have other evaluation metrics- accuracy and outlier which are defined earlier.
The lower the MAE/residual, the more precise the
model is. The accuracy metrics give us the statistics about how many samples are with an acceptable level of accuracy. The standard deviation estimates the average proportion of the variation in the response
around the mean. The outlier in prediction provides insight into how many large error samples are in the prediction, which gives a better  picture of the distribution of deviation.   
Table \ref{tab:space} shows the performance metrics on all three trained models on the test data set. All three models are trained on the same $8000$ training data set and tested on the $3311$ test data set. Our DNN model performed at significantly higher accuracy (92\%) in comparison to other models - Random forest (72\%) and  Gradient boost regressor (47 \%). Our trained model also has one-third less number of outliers than the Random forest and Gradient Boost models. In all the metrics our model performs better than  Random forest and Gradient Boost based regressors.
 \begin{table}
\captionsetup{justification=centering,labelsep=period}
\caption{Key performance metrics }
\label{tab:space}
\begin{tabular}{|c|c|c|c|c|} 
\hline
\textbf{Model} & \textbf{Accuracy} & \textbf{Residual} &\textbf{Outlier} &\textbf{Deviation}\\
\hline
Deep ensemble & 92.20\% & 0.045  & 48 & 118.67 \\
\hline
Random Forest & 72.27\% & 0.13 & 164 & 141.23  \\
\hline
Gradient Boost & 47.47\% &  0.21 &  157 & 132.60\\
\hline
\end{tabular}
\end{table}

\section{Related Work}
\label{sec:related_work}

AI has the ability to automatically extract features and learn a surrogate model using training data. Its applications are extensive ranging from the operation of systems\cite{vardhan2021rare}\cite{abbeel2010autonomous} to the design of systems\cite{vardhan2021machine}. The FEA-based surrogate is nowadays gaining popularity and has application in biomechanical problems, like stress analysis of the aorta \cite{liang2018deep}\cite{madani2019bridging}, 
and the biomechanical behavior of human soft tissue \cite{martinez2017finite}.  Liang et al. \cite{liang2018deep} developed patient-specific models of the stress distribution of the
aorta using as input the FEA data and directly
outputting the stress distribution. Madani et al.\cite{madani2019bridging} developed an FEA-based surrogate model to
predict real-time the maximum von Mises stress magnitude with an average error of less than 10\%. They could estimate the  aortic wall stress distribution in patients affected
by atherosclerosis without the need of highly trained specialists, therefore allowing quick
clinical analysis of the patient health.

\section{Conclusion}
\label{sec:conclusion}
In this work, we showed that by using AI and especially deep learning, we can approximate the FEA behavior with great accuracy on a large class of pressure vessel design problems and can bypass the computationally costly Finite element analysis-based numerical methods. This dataset and model can be used for initial design evaluation for real-world applications.  The trained model and the dataset is available to public access and can be used by sub-sea pressure vessel designer for finding initial good designs. We also showed that our trained model has much superior accuracy than other regression models. 
Since the average simulation time per simulation is 3 minutes, we are still generating some more data and the future extension of this work would be adding some more data to make this model much more accurate. Another future work is integrating this surrogate model with optimization tools and bypassing the FEA numerical simulator from the loop. The expected benefit would be almost no optimization time for a requirement.  These designs can be again verified with FEA based numerical process. 

\section{Acknowledgments}

This work is supported by DARPA’s Symbiotic Design for CPS project and by the Air Force Research Laboratory (FA8750-20-C-0537).

\bibliographystyle{IEEEtran}
\bibliography{bibliography}

\begin{thebibliography}{10}
\providecommand{\url}[1]{#1}
\csname url@samestyle\endcsname
\providecommand{\newblock}{\relax}
\providecommand{\bibinfo}[2]{#2}
\providecommand{\BIBentrySTDinterwordspacing}{\spaceskip=0pt\relax}
\providecommand{\BIBentryALTinterwordstretchfactor}{4}
\providecommand{\BIBentryALTinterwordspacing}{\spaceskip=\fontdimen2\font plus
\BIBentryALTinterwordstretchfactor\fontdimen3\font minus
  \fontdimen4\font\relax}
\providecommand{\BIBforeignlanguage}[2]{{%
\expandafter\ifx\csname l@#1\endcsname\relax
\typeout{** WARNING: IEEEtran.bst: No hyphenation pattern has been}%
\typeout{** loaded for the language `#1'. Using the pattern for}%
\typeout{** the default language instead.}%
\else
\language=\csname l@#1\endcsname
\fi
#2}}
\providecommand{\BIBdecl}{\relax}
\BIBdecl

\bibitem{raissi2019physics}
M.~Raissi, P.~Perdikaris, and G.~E. Karniadakis, ``Physics-informed neural
  networks: A deep learning framework for solving forward and inverse problems
  involving nonlinear partial differential equations,'' \emph{Journal of
  Computational physics}, vol. 378, pp. 686--707, 2019.

\bibitem{vardhan2019modeling}
H.~Vardhan, N.~M. Sarkar, and H.~Neema, ``Modeling and optimization of a
  longitudinally-distributed global solar grid,'' in \emph{2019 8th
  International Conference on Power Systems (ICPS)}.\hskip 1em plus 0.5em minus
  0.4em\relax IEEE, 2019, pp. 1--6.

\bibitem{xu2015empirical}
B.~Xu, N.~Wang, T.~Chen, and M.~Li, ``Empirical evaluation of rectified
  activations in convolutional network,'' \emph{arXiv preprint
  arXiv:1505.00853}, 2015.

\bibitem{he2016deep}
K.~He, X.~Zhang, S.~Ren, and J.~Sun, ``Deep residual learning for image
  recognition,'' in \emph{Proceedings of the IEEE conference on computer vision
  and pattern recognition}, 2016, pp. 770--778.

\bibitem{sagi2018ensemble}
O.~Sagi and L.~Rokach, ``Ensemble learning: A survey,'' \emph{Wiley
  Interdisciplinary Reviews: Data Mining and Knowledge Discovery}, vol.~8,
  no.~4, p. e1249, 2018.

\bibitem{kingma2014adam}
D.~P. Kingma and J.~Ba, ``Adam: A method for stochastic optimization,''
  \emph{arXiv preprint arXiv:1412.6980}, 2014.

\bibitem{breiman2017classification}
L.~Breiman, J.~H. Friedman, R.~A. Olshen, and C.~J. Stone, \emph{Classification
  and regression trees}.\hskip 1em plus 0.5em minus 0.4em\relax Routledge,
  2017.

\bibitem{vardhan2021rare}
H.~Vardhan and J.~Sztipanovits, ``Rare event failure test case generation in
  learning-enabled-controllers,'' in \emph{2021 6th International Conference on
  Machine Learning Technologies}, 2021, pp. 34--40.

\bibitem{abbeel2010autonomous}
P.~Abbeel, A.~Coates, and A.~Y. Ng, ``Autonomous helicopter aerobatics through
  apprenticeship learning,'' \emph{The International Journal of Robotics
  Research}, vol.~29, no.~13, pp. 1608--1639, 2010.

\bibitem{vardhan2021machine}
H.~Vardhan, P.~Volgyesi, and J.~Sztipanovits, ``Machine learning assisted
  propeller design,'' in \emph{Proceedings of the ACM/IEEE 12th International
  Conference on Cyber-Physical Systems}, 2021, pp. 227--228.

\bibitem{liang2018deep}
L.~Liang, M.~Liu, C.~Martin, and W.~Sun, ``A deep learning approach to estimate
  stress distribution: a fast and accurate surrogate of finite-element
  analysis,'' \emph{Journal of The Royal Society Interface}, vol.~15, no. 138,
  p. 20170844, 2018.

\bibitem{madani2019bridging}
A.~Madani, A.~Bakhaty, J.~Kim, Y.~Mubarak, and M.~R. Mofrad, ``Bridging finite
  element and machine learning modeling: stress prediction of arterial walls in
  atherosclerosis,'' \emph{Journal of biomechanical engineering}, vol. 141,
  no.~8, 2019.

\bibitem{martinez2017finite}
F.~Mart{\'\i}nez-Mart{\'\i}nez, M.~J. Rup{\'e}rez-Moreno,
  M.~Mart{\'\i}nez-Sober, J.~A. Solves-Llorens, D.~Lorente,
  A.~Serrano-L{\'o}pez, S.~Mart{\'\i}nez-Sanchis, C.~Monserrat, and J.~D.
  Mart{\'\i}n-Guerrero, ``A finite element-based machine learning approach for
  modeling the mechanical behavior of the breast tissues under compression in
  real-time,'' \emph{Computers in biology and medicine}, vol.~90, pp. 116--124,
  2017.

\end{thebibliography}

\end{document}